\documentclass[11pt]{article}

\usepackage[final]{acl}

\usepackage{times}
\usepackage{latexsym}

\usepackage[T1]{fontenc}

\usepackage[utf8]{inputenc}

\usepackage{microtype}

\usepackage{inconsolata}

\usepackage{ifthen}
\usepackage{graphicx}
\usepackage{multirow}
\usepackage{todonotes}
\usepackage{amsmath}
\usepackage{subcaption}
\usepackage{cleveref}
\usepackage{booktabs}
\usepackage{listings}
\usepackage{siunitx}

\usepackage{tikz}
\usetikzlibrary{shapes.geometric, shapes.misc, arrows.meta, positioning}

\usepackage{tcolorbox}
\tcbuselibrary{breakable, skins}

\newtcolorbox[auto counter, number within=section]{promptbox}[2][]{
  title={Prompt \thetcbcounter: #2},
  colback=gray!8,
  colframe=black!60,
  fonttitle=\bfseries\small,
  breakable,
  #1
}

\title{Losses that Cook: Topological Optimal Transport \\ for Structured Recipe Generation}

\author{
 \textbf{Mattia Ottoborgo\textsuperscript{1}},
 \textbf{Daniele Rege Cambrin\textsuperscript{2}},
 \textbf{Paolo Garza\textsuperscript{2}}
 \\
 \textsuperscript{1}Trustpilot \hspace{5mm}
 \textsuperscript{2}Politecnico di Torino
 \\
 \small{
   \textbf{Correspondence:} maot@trustpilot.com,\{daniele.regecambrin,paolo.garza\}@polito.it
 }
}

\begin{document}
\maketitle
\begin{abstract}
Cooking recipes are complex procedures that require not only a fluent and factual text, but also accurate timing, temperature, and procedural coherence, as well as the correct composition of ingredients. Standard training procedures are primarily based on cross-entropy and focus solely on fluency. Building on RECIPE-NLG, we investigate the use of several composite objectives and present a new topological loss that represents ingredient lists as point clouds in embedding space, minimizing the divergence between predicted and gold ingredients. Using both standard language generation metrics and recipe-specific metrics, we find that our loss significantly improves ingredient- and action-level metrics. Meanwhile, the Dice loss excels in time/temperature precision, and the mixed loss yields competitive trade-offs with synergistic gains in quantity and time. A human preference analysis supports our finding, showing our model is preferred in 62\% of the cases. 
\end{abstract}

\section{Introduction}
Generating usable cooking recipes with language models requires more than fluent text: models must produce ingredients, quantities, and step-by-step instructions that are factually correct, numerically plausible, and procedurally executable. In this setting, errors on a few key tokens (e.g., omitting “eggs” in carbonara pasta or doubling the cooking temperature) can render the entire recipe unusable, even if the text is fluent and the semantics are similar to the correct output \cite{bien2020,guoshan2025}.

Standard fine-tuning with cross-entropy (CE) is ill-suited to this challenge because it treats all tokens as equally important, despite a strong asymmetry between high-impact tokens (ingredients, quantities, times, temperatures, core actions) and low-impact connective words \cite{Chen2024}. This misalignment manifests in common failure modes: poor ingredient recall, inaccurate quantities, and instruction sequences that are syntactically plausible but procedurally incorrect. Existing work on recipe generation and structured text generation has largely relied on CE-based objectives, beam search, or schema-constrained decoding, but has not directly targeted the holistic composition of ingredient sets and numerical aspects of recipes through the training loss \cite{Lam2024}.

In parallel, some works have explored alternative or auxiliary objectives in other natural language processing (NLP) tasks, such as focal and dice losses \cite{regecambrin2024beyond}. These approaches suggest that rethinking the loss can steer models toward rare but important events or holistic set properties; however, they have not been systematically applied to structured recipe generation and do not exploit the inherent topology of recipes. Moreover, standard natural language generation (NLG) metrics, such as ROUGE \cite{lin2004} and BERTScore \cite{Zhang2020}, capture fluency and semantic similarity but fail to directly measure whether a recipe accurately specifies the ingredients, quantities, and cooking parameters.

This work addresses these gaps by focusing on Small Language Models (SLMs) fine-tuned for structured recipe generation on a focused subset of the RECIPE-NLG corpus \cite{bien2020} (pasta, rice, and sandwiches). We introduce a topological loss that represents ingredient lists as point clouds in embedding space and minimizes a Sinkhorn divergence \cite{cuturi2013} between predicted and gold ingredients, explicitly encoding ingredient-level structure beyond token-wise CE. We further investigate how combining our proposal with existing losses can balance ingredient structure with numerical accuracy. To evaluate these objectives, we design a recipe-specific metric suite, including ingredient recall, quantity precision, action and step edit distances, and time/temperature precision, in addition to standard text metrics.

Our experiments demonstrate that augmenting CE with the proposed topological loss substantially improves ingredient recall, quantity precision, and procedural accuracy over CE alone, while Dice-based losses excel in terms of time and temperature precision. The mixed loss yields well-rounded trade-offs and, in some cases, synergistic gains (e.g., in quantity and time precision), with many improvements over CE and single custom losses. Together, these results demonstrate that carefully designed loss functions can meaningfully improve structured recipe generation in SLMs without increasing model size or inference-time complexity. We release the code for reproducibility on 
\makeatletter
\ifacl@finalcopy
\url{https://github.com/DarthReca/losses-cook}.
\else
\url{https://anonymous.4open.science/r/losses-cook-C08B/}.
\fi
\makeatother

\section{Methodology}
This section presents the task, the dataset, the proposed loss, and the evaluation metrics.

\subsection{Task}
We formalize structured recipe generation as a mapping $f: P_{in} \rightarrow R_{out}$, where $P_{in}$ is a natural language prompt (e.g., "Generate a recipe for Pasta Carbonara") and $R_{out} = {I, S}$ is a structured JavaScript Object Notation (JSON) output containing: (1) a list of ingredients $I = {I_1, I_2, \ldots, I_n}$, and (2) a list of instruction steps $S = {S_1, S_2, \ldots, S_m}$ as shown in \Cref{sec:dataset_details}. The objective is to learn $f$ that satisfies multiple constraints simultaneously: fluency, factual correctness (appropriate ingredients), numerical accuracy (plausible quantities, times, temperatures), and procedural coherence (logical instruction sequences).

\subsection{Dataset}
We use a subset of 5,000 recipes from RECIPE-NLG \cite{bien2020}, focusing on pasta, rice, and sandwich dishes to ensure distributional consistency between the training and test sets (knowing how to cook rice does not necessarily enable you to prepare a good steak). To improve domain-specific learning, we augment the dataset with 235 manually curated cooking questions covering: missing ingredient identification, ingredient substitution, recipe scaling, quantity reasoning, time estimation, and temperature specification. These questions teach the model critical relationships between ingredients and numerical reasoning required for recipe generation. More details and examples are reported in \Cref{sec:dataset_details}.

\subsection{Loss Functions}
Standard Cross-Entropy (CE) minimizes $\mathcal{L}_{\text{CE}} = -\log p_c$, where $p_c$ is the model's predicted probability for the correct token $c$. We establish CE-only fine-tuning as our primary baseline. We also evaluate Focal Loss \cite{tsung2017}, which down-weights easy examples via a modulating factor $(1-p_t)^\gamma$ to address token frequency imbalance, and Dice Loss \cite{Sudre2017}, which optimizes set-level overlap using a differentiable Dice coefficient to encourage correct token sets.

\subsubsection{Topological Loss}
Our main contribution is a topological loss that operates in embedding space to capture the structural coherence of the ingredient section. The key insight is that token sequences representing semantically similar ingredient lists should form geometrically similar shapes in embedding space, unlike cross-entropy, which treats all substitutions equally regardless of semantic proximity.

We construct two point clouds in embedding space from all tokens within the ingredient section: (1) For the predicted recipe, we generate soft probabilistic embeddings by applying softmax to output logits $P = softmax(logits)$ and computing weighted embedding averages $emb_{soft} = P \cdot E$, where $E$ is the model's token embedding matrix as shown in \Cref{fig:soft_embeddings}. (2) For the ground truth, we perform standard embedding lookups. 

\begin{figure}[htb]
    \centering
    \resizebox{\linewidth}{!}{\usetikzlibrary{shapes.geometric, arrows.meta, positioning}

\begin{tikzpicture}[
    node distance=1.3cm and 3.5cm,
    box/.style   = {rectangle, rounded corners=4pt, draw=#1!60!black,
                    fill=#1!15, minimum width=3.2cm, minimum height=0.85cm,
                    font=\small, align=center},
    llm/.style   = {trapezium, trapezium left angle=75, trapezium right angle=105,
                    draw=yellow!60!black, fill=yellow!20,
                    minimum width=3.2cm, minimum height=0.85cm,
                    font=\small\bfseries},
    diam/.style  = {diamond, draw=yellow!60!black, fill=yellow!20,
                    aspect=2.8, minimum width=3.4cm, font=\small\bfseries},
    emb/.style   = {rectangle, rounded corners=4pt, draw=blue!50!black,
                    fill=blue!10, minimum width=3.2cm, font=\small, align=center},
    arr/.style   = {->, >=Stealth, thick},
    darr/.style  = {->, >=Stealth, thick, dashed, gray}
]

\node[box=yellow]           (prompt)  {Prompt: Carbonara Recipe};
\node[llm,  below=of prompt](llm)     {LLM};
\node[box=yellow,below=of llm](softmax){Softmax};
\node[diam, below=of softmax](ws)     {Weighted Sum};
\node[box=blue, below=of ws](out)     {$\boldsymbol{emb_{soft}}$};

\node[emb, right=1cm of ws] (emb) {
    \textbf{Embedding Matrix E}\\
    \rule{2.8cm}{0.4pt}\\       %
    pasta \\ egg \\ guanciale \\ $\cdots$
};

\draw[arr] (prompt)  -- (llm);
\draw[arr] (llm)     -- node[right=2pt, font=\footnotesize\itshape] {Token Logits $z$} (softmax);
\draw[arr] (softmax) -- node[right=2pt, font=\footnotesize\itshape] {Probabilities $p$} (ws);
\draw[arr] (emb)     -- (ws);
\draw[arr] (ws)      -- (out);

\draw[darr] (llm.east) to[out=0, in=90]
    node[above, sloped, font=\footnotesize\itshape, gray] {weights access}
    (emb.north);

\end{tikzpicture}}
    \caption{Soft embedding computation: output logits $z$ are converted into token probabilities $p$ using softmax, which are used to compute a differentiable weighted average over the model's embedding matrix $E$ to create $emb_{soft} = p \cdot E$.}
    \label{fig:soft_embeddings}
\end{figure}

\begin{figure}[htb]
    \centering
    \includegraphics[width=0.8\linewidth]{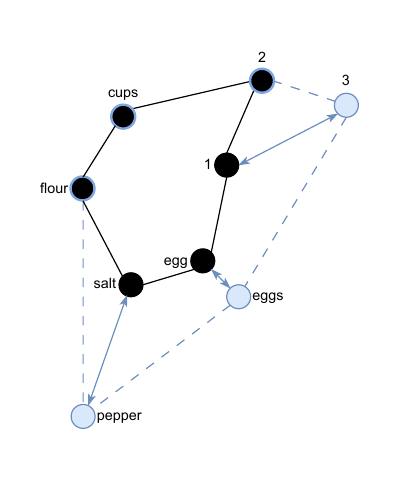}
    \caption{The loss aligns the ground truth (black) and predicted (blue) token in embedding space. Shared tokens like "flour" (black dots with blue halos) have zero transport cost. The loss minimizes the transport distance for divergent tokens, penalizing semantic shifts (e.g., "salt" $\to$ "pepper") and structural deviations (e.g., "egg" $\to$ "eggs")}
    \label{fig:point_clouds}
\end{figure}

The loss then measures the geometric dissimilarity between these clouds using Sinkhorn divergence $\mathcal{S}_\epsilon$ \cite{cuturi2013,Cuturi2016}, a differentiable approximation of optimal transport (Wasserstein) distance:

$$\mathcal{L}_{\text{Topo}} = \mathcal{S}_\epsilon (PC_{pred}, PC_{target})$$

where $\epsilon$ is the entropic regularization parameter. This encourages the model to generate ingredient lists that are both semantically and structurally coherent in the embedding space, not just token-wise accurate, as shown in \Cref{fig:point_clouds}.

\subsection{Evaluation Metrics}
To comprehensively assess recipe quality, we combine standard NLG metrics with recipe-specific measures. We report ROUGE-1 (R1) and BERTScore F1 (BS) to measure linguistic fluency and semantic coherence. Since factual and procedural correctness are paramount, we introduce ad-hoc metrics. Ingredient Recall (IR) is the fraction of ground-truth ingredients correctly generated; Quantity Precision (QP) is the accuracy of numerical quantities for correctly recalled ingredients; Action Precision (AP) is the precision of key cooking verbs (e.g., boil, fry, sauté) in generated instructions; Action (AD) and Step (SD) Edit Distances are Levenshtein distance between sequences of cooking actions or full instruction steps with times/temperatures, measuring procedural correctness; Time (TiP) and Temperature (TeP) Precision are the accuracy of time durations and temperatures mentioned in instructions. Additional details on the metrics computation are provided in \Cref{sec:extraction_pipeline}.

\section{Experiments}
In this section, we present the experimental settings and results.

\subsection{Experimental Settings}
We fine-tune a pre-trained Qwen3-4B \cite{yang2025} model with Low-Rank Adaptation (LoRA) \cite{hu2022} and the AdamW optimizer. More training details are provided in \Cref{sec:experimental_settings}. All custom losses (i.e., dice, focal, and topological) are combined with cross-entropy (CE) as a composite objective to maintain linguistic fluency while enhancing domain-specific correctness. In all composite setups with a single custom loss, the objective is $L = 0.6L_{CE} + 0.4L_{custom}$, chosen empirically to preserve fluency while amplifying the signal on critical tokens. We also train a mixed-loss configuration with $L = 0.6L_{CE} + 0.2L_{Dice} + 0.2L_{Topo}$.  The same augmented dataset is used for all fine-tuning conditions. We compared our model against a commercial model (Gemini 2.0 Flash) and a larger version of Qwen3 with 14B parameters. To assess 
whether the observed trends generalize across architectures and parameter scales, we additionally evaluate SmolLM3-3B and Qwen2.5-1.5B; those results are reported in \Cref{sec:additional_results}.

\subsection{Quantitative Evaluation}

\begin{table*}[htb]
  \sisetup{table-format=3.2, round-precision=2, round-mode=places, detect-all=true}
  \centering
  \resizebox{\linewidth}{!}{
      \begin{tabular}{c | l |S S| S S S S S S S}
        \toprule
        & Model         & {R1 $\uparrow$} & {BS $\uparrow$} & {AP $\uparrow$} & {QP $\uparrow$} & {IR $\uparrow$} & {TeP $\uparrow$} & {TiP $\uparrow$} & {AD $\downarrow$} & {SD $\downarrow$} \\ \midrule
        \multirow{3}{*}{\rotatebox{90}{No-FT}}
        & Gemini 2.0 & 15.0847    & 88.50          & 43.80          & 44.51          & 37.47          & \itshape 76.88           & 36.92           & 36.21            & 48.60 \\
        & Qwen3-14B     & 25.2393       & 85.69          & 42.12          & 44.51          & 28.25          & \bfseries 81.71           & 41.98           & 35.97            & 38.09            \\
        & Qwen3-4B    & 22.4946       & 87.93          & 32.40          & 25.09          & 26.09          & 39.84           & 41.45           & 48.50            & 52.09            \\ \midrule
        \multirow{5}{*}{\rotatebox[origin=c]{90}{Qwen3-4B FT}}
        & CE            & 27.3048       & 88.78          & 45.09          & 50.94          & 35.98          & 61.93           & 52.09           & 37.83            & 39.48            \\
        & Focal         & 26.0949       & 89.94          & 41.09          & 54.94          & 43.09          & 59.60           & 48.52           & 37.40            & 38.05            \\
        & Dice          & 29.8744       & 90.49          & 50.59          & 57.44          & 44.90          & 74.58           & \itshape 59.68           & \itshape 31.09            & \itshape 35.08            \\
        & Topological   & \itshape 30.3985       & \itshape 90.97          & \bfseries 59.68          & \itshape 63.93          & \bfseries 48.59          & 65.59           & 55.55           & \bfseries 30.49            & \bfseries 34.09            \\ \cmidrule{2-11}
        & Topo+Dice     & \bfseries 31.9045     & \bfseries 90.99          & \itshape 57.59          & \bfseries 65.09          & \itshape 47.09          & 67.89           & \bfseries 61.95           & \bfseries 30.49            & \bfseries 34.09            \\              \bottomrule
    \end{tabular}
  }
  \caption{Results for finetuned Qwen3-4B and pretrained models using ROUGE-1 (R1), BERTScore (BS), Action Precision (AP), Quantity Precision (QP), Ingredient Recall (IR), Temperature Precision (TeP), Time Precision (TiP), Action Distance (AD), Step Distance (SD). In \textbf{bold} the top performance, and in \textit{italic} the second best. FT = FineTuned; No-FT = Non FineTuned. Results on additional architectures (SmolLM3-3B, Qwen2.5-1.5B) are reported in \Cref{sec:additional_results}.}
  \label{tab:results}
\end{table*}

As shown in \Cref{tab:results}, strong pretrained instruction-tuned LLMs (Gemini 2.0, Qwen3-14B, and Qwen3-4B) underperform our fine-tuned models on both general NLG metrics (R1, BS) and on recipe-specific measures. While Qwen3-14B improves over Gemini in R1 and yields the best temperature precision (TeP), both models exhibit substantially weaker action/ingredient grounding (e.g., lower AP and IR) and larger procedural divergences (AD, SD) than the fine-tuned objectives. This suggests that general conversational competence does not directly translate into executable, constraint-satisfying recipe generation, highlighting the importance of domain adaptation for maintaining ingredient coverage and step-level alignment.

Across fine-tuned configurations, even pure cross-entropy (CE) yields large improvements over the pretrained baselines in AP, QP, IR, TiP, and TeP, while also reducing AD and SD, indicating better procedural faithfulness to the ground-truth recipes. Among composite objectives, focal loss slightly improves BS and IR over CE, but lags behind Dice and Topological losses on most task-specific metrics, indicating that reweighting difficult tokens alone is insufficient to enforce fine-grained culinary constraints (quantities, times, and action sequences). Consistent with this, our significance tests indicate that improvements in BS are not always reliable, reinforcing that the primary benefits of the custom objectives are manifested in structure- and constraint-sensitive metrics rather than generic semantic similarity.

Dice loss yields stronger gains than CE and focal in both numerical and procedural dimensions, with notable improvements in QP, TeP, and TiP, alongside lower edit distances, underscoring its effectiveness in covering critical tokens and reducing omissions. Our topological loss delivers the most consistent improvements over CE across the recipe-specific metrics, substantially increasing AP, QP, and IR, while achieving the lowest (or tied-lowest) AD and SD. This suggests that aligning ingredient point clouds in embedding space facilitates coherent ingredient sets and step sequences, surpassing what token-wise objectives can capture. The mixed Topo+Dice objective further improves overall performance, achieving the best R1, BS, QP, and TiP while matching the best edit distances, suggesting that Dice and topological alignment are complementary—Dice sharpens numerical/procedural accuracy, whereas the topological term enforces ingredient-level semantic structure. These trends are consistent across architectures and parameter scales, as confirmed by experiments on SmolLM3-3B and Qwen2.5-1.5B reported in \Cref{sec:additional_results}.

\subsection{Human Evaluation}
We conducted pairwise comparisons on 64 recipes (balanced across pasta, rice, and sandwiches) with 3 expert annotators evaluating ingredients, numbers, procedure, and overall quality. Inter-annotator agreement was substantial (mean Fleiss' $\kappa = 0.69$) with full (3/3) and partial (2/3) agreement in 71\% and 28\% of cases. Topo+Dice was preferred over CE for ingredients (38\% vs 10\%), procedure (46\% vs 9\%), numbers (42\% vs 9\%), and overall quality (62\% vs 11\%), corresponding to win rates of 79\%, 81\%, 88\%, and 84\% when excluding ties. All differences are statistically significant ($p < 0.01$, $\chi^2$, and McNemar tests). Error analysis revealed a 67.5\% reduction in generation errors (7.8\% vs 24.0\%), with Topo+Dice virtually eliminating critical errors such as step mismatches and hallucinations. More details are provided in \Cref{sec:human_evaluation_details}.

\section{Conclusion}
We studied structured recipe generation where models must jointly satisfy fluency, plausibility, and coherence. To address a key weakness of token-level objectives, we introduced a topological loss that aligns predicted and reference ingredient sets as point clouds in embedding space. This improves over CE baselines, and it complements Dice loss in a mixed objective that yields the best overall balance between linguistic quality and constraint satisfaction. Future work includes scaling to broader cuisines and dietary constraints, as well as incorporating explicit safety- and nutrition-aware validation to better support real-world use.

\makeatletter
\ifacl@finalcopy
\section*{Acknowledgments}
The authors would like to thank Sebastiano Bonfanti and Luca Barbotti for their valuable contribution to the manual evaluation process.
\fi
\makeatother

\section*{Limitations}
This work has several limitations. The training and evaluation domain is intentionally narrow (pasta, rice, and sandwiches), so the observed improvements may not transfer to other cuisines, cooking styles, or multi-course recipes without additional adaptation.
Second, our curated question augmentation is relatively small and manually constructed; it may not cover the full diversity of quantity, time, and temperature reasoning failures encountered in open-ended prompting. Third, several proposed evaluation measures rely on automated extraction of ingredients, quantities, actions, and temperatures from generated text; these pipelines can be brittle to paraphrase, formatting deviations, or uncommon culinary terminology, which can introduce noise into metric estimates.
Fourth, the topological loss depends on the geometry of the model's embedding space and hyperparameters, and it introduces computational overhead relative to CE-only fine-tuning. Fifth, the human evaluation is small-scale and based on convenience sampling, which limits statistical power and generalizability. Fifth, the method compares entire token sequences rather than individual ingredient entities, meaning it captures sequence-level geometric similarity but does not explicitly parse or compare individual ingredients. Finally, while we measure some safety-related aspects (e.g., temperatures and times), the system is not guaranteed to produce safe recipes, respect allergies, or provide nutritionally appropriate guidance, and should not be treated as a substitute for professional advice.

\section*{Ethical Considerations}
This paper focuses on improving factual and procedural correctness in recipe generation, but generated recipes can still cause harm if followed blindly. Potential risks include incorrect or missing allergen-related ingredients, unsafe handling suggestions, and implausible times/temperatures that could lead to undercooking or other food safety issues.
To mitigate misuse, models and demos built on this work should include clear user-facing warnings, encourage verification against trusted culinary sources, and, where possible, apply post-generation validation (e.g., allergy constraints, range checks for temperatures and times, and consistency checks between ingredients and steps).

Regarding data and bias, recipes reflect cultural preferences and may over-represent certain cuisines, ingredients, or cooking norms, which can lead to uneven performance across user groups and prompts.
The dataset was used for research on recipe text generation; nevertheless, users should be aware that generated outputs may reproduce biases present in the training data (e.g., assumptions about dietary norms).

For human evaluation, participants were adult volunteers recruited via convenience sampling, participated without compensation, and could stop at any time. No sensitive personal data was collected beyond basic background relevant to the evaluation task.

\bibliography{custom}

\appendix

\section{Dataset Details}
\label{sec:dataset_details}

This section provides additional details on the training dataset.

\subsection{Dataset Preparation}
Recipe-NLG contains recipes expressed in a convenient format, as they already include two fields called “ingredients” and “instructions” that contain a list of the respective entities. The primary aspect that needed modification for our experiment was the conversion of the unit of measure for the ingredients and other entities to the metric system. We decided to convert to the metric system to ensure reliable and consistent numerical values for ingredient quantities, time, and temperature. To achieve this, we utilized Gemini 2.0 Flash \cite{Hassabis2024Gemini} to convert all quantities to the metric system. The full conversion prompt used with LangExtract \footnote{\url{https://langextract.com/}} is reported in Prompt \ref{prompt:metric}. An example transformation can be seen in the \Cref{fig:metric_example}.

\begin{promptbox}[label={prompt:metric}]{Imperial-to-Metric Conversion}
\textbf{System:} You are a precision data extraction tool specializing in culinary text.
Your task is to analyze a list of ingredients and extract specific entities into a
structured JSON format.

\medskip
\textbf{Instructions:} For each ingredient in the input list, extract three fields:
\texttt{quantity}, \texttt{unit}, and \texttt{ingredient}.

\begin{itemize}
  \item \textbf{Quantity:} Extract the numerical value. Convert fractions to decimals.
        If no quantity is specified, assume 1.
  \item \textbf{Unit:} Must be one of \texttt{"g"}, \texttt{"ml"}, or \texttt{"unit"}.
        Convert all imperial and household measurements (e.g., spoons, cups, cans) to
        their estimated metric equivalent. Use grams for solids, ml for liquids, and
        unit for countable items (e.g., eggs).
  \item \textbf{Ingredient:} Extract only the core base name.
\end{itemize}

\medskip
\textbf{Example:}\\
Input: \texttt{"3/4 c.\ chopped onion"} \\
Output: \texttt{\{"quantity": 180, "unit": "ml", "ingredient": "onion"\}}
\end{promptbox}

\lstset{
  basicstyle=\ttfamily\scriptsize,
  backgroundcolor=\color{gray!8},
  frame=single,
  breaklines=true,
  columns=flexible,
  xleftmargin=4pt,
  xrightmargin=4pt,
}

\begin{figure}[htb]
\centering
\begin{minipage}[t]{0.47\linewidth}
\lstset{title={\small\textbf{Before (Imperial)}}}
\begin{lstlisting}
{
  "ingredients": [
    "10.5 oz spaghetti",
    "5.3 oz guanciale, diced",
    "2 large eggs",
    "1.8 oz Pecorino Romano, grated",
    "0.9 oz Parmigiano, grated",
    "Black pepper, to taste"
  ],
  "instructions": [ ... ]
}
\end{lstlisting}
\end{minipage}
\hfill
\begin{minipage}[t]{0.47\linewidth}
\lstset{title={\small\textbf{After (Metric)}}}
\begin{lstlisting}
{
  "ingredients": [
    "300g spaghetti",
    "150g guanciale, diced",
    "2 large eggs",
    "50g Pecorino Romano, grated",
    "25g Parmigiano, grated",
    "Black pepper, to taste"
  ],
  "instructions": [ ... ]
}
\end{lstlisting}
\end{minipage}
\caption{Example of imperial-to-metric conversion applied to a carbonara recipe. Instructions omitted for brevity.}
\label{fig:metric_example}
\end{figure}

\subsection{Question Examples}
\label{sec:dataset_examples}
In addition to the recipe collection, we manually curated and added a new dataset of 235 questions related to general knowledge in the cooking domain. We decided to augment the dataset with this set of questions to enable the model not only to learn how to write recipes in general, but also to understand how to better handle quantities, ingredient similarities and differences, ingredient proportions, and other challenges related to recipe generation. The questions have been generated from a set of humanly written samples that have been augmented using Gemini 2.0 Flash \cite{Hassabis2024Gemini}. The resulting augmented set has been then evaluated by humans for the quality of questions, diversity in recipes and ingredients involved, and potential data leaks in relation to the test dataset. The questions can be categorized into the following:

\paragraph{Missing ingredient identification.} It consists of presenting recipes along with an incomplete list of ingredients. The questions, therefore, evaluate the model's capacity to identify the missing ingredients and learn the relationships between them.

\begin{quote}
    \textbf{Question}: You are an expert chef. You have been tasked with generating a recipe for Carbonara. The following ingredients have been identified: pasta, pecorino, and eggs. Which ingredient(s) are missing?
    
    \textbf{Answer}: pepper, guanciale
\end{quote}

\paragraph{Substitution validation.}
This teaches the model about similarity and differences between ingredients. Our goal is to teach the model to recognize similar candidates.

\begin{quote}
    \textbf{Question}: You are an expert chef. You have been tasked with generating a recipe for Carbonara. Pecorino is not available, and you have been asked to propose an alternative ingredient. Which ingredient can best substitute it?
    
    \textbf{Answer}: Parmesan
\end{quote}

\paragraph{Recipe scaling.}
The goal is to identify the scaling rules used for different types of ingredients when adapting the recipe to serve a different number of people.

\begin{quote}
    \textbf{Question}: You are an expert chef. You have been tasked with modifying the following list of ingredients, intended for two servings, to accommodate 4 servings. Modify the quantities of each ingredient. List of ingredients: 200g of pasta, 100g of pecorino, 3 yolks, 140g of guanciale.
    
    \textbf{Answer}: 400g of pasta, 100g of pecorino, 3 yolks, 140g of guanciale.
\end{quote}

\paragraph{Quantity ingredient.}
They aim to teach the model how to properly dose ingredients based on the context.

\begin{quote}
    \textbf{Question}: You are an expert chef. How many grams of spaghetti are needed to prepare a Carbonara for two people?
    
    \textbf{Answer}: 400g
\end{quote}

\paragraph{Time}
This set of questions aims to teach the model how to properly define times in relation to the different steps involved.

\begin{quote}
    \textbf{Question}: You are an expert chef. How much time do you need to boil pasta?
    
    \textbf{Answer}: 10 minutes.
\end{quote}

\paragraph{Temperature}
This set of questions aims to teach the optimal temperature at which certain ingredients should be exposed to accomplish a specific step.

\begin{quote}
    \textbf{Question}: You are an expert chef. At which temperature should you boil pasta?
    
    \textbf{Answer}: 100 Celsius degrees.
\end{quote}

\section{Extraction Pipeline}
\label{sec:extraction_pipeline}
All models generate 1,000 recipes in JSON format with schema enforcement via the Outlines library \cite{willard2023} to prevent malformed outputs. We then apply NER using a Qwen3-4B to extract quantities, units, times, temperatures, and primary actions from natural language instructions, enabling precise metric computation using LangExtract \footnote{\url{https://langextract.com/}} with the Prompts \ref{prompt:ner_directions} and \ref{prompt:metric}.

\begin{promptbox}[label={prompt:ner_directions}]{Directions Extraction}
\textbf{System: }You are an expert culinary data scientist. Your task is to analyze a list of cooking recipe steps and extract key information for each step into a structured JSON format.

\textbf{Instructions:} Process the input list of recipe steps in the order they are provided. For each step, extract the following four entities:
\begin{itemize}
    \item action: The single, primary cooking verb for the step (e.g., "Combine", "Bake", "Stir").
    \item temperature\_celsius: The temperature mentioned, converted to Celsius. If not present, use null. Must be a number.
    \item time\_minutes: The duration in minutes. If a range is given (e.g., "10-12 minutes"), use the average. If not present, use null. Must be number
    \item ingredients: A list of the food ingredients involved in that specific step. If no ingredients are mentioned, use an empty list [].
\end{itemize}
\end{promptbox}

\section{Additional Experimental Settings}
\label{sec:experimental_settings}
We used LoRa with bfloat16 training, and the learning rate is $10^{-4}$ for 2 epochs. The batch size is 2 with gradient accumulation of 3. The Sinkhorn $\epsilon$ parameter is set to 0.05, and the focal loss $\gamma$ is set to 2. The \Cref{tab:lora_hyperparams} summarizes LoRa hyperparameters, while the targeted modules are q\_proj, k\_proj, v\_proj, o\_proj.

\begin{table}[h]
  \centering
  \begin{tabular}{lc}
    \toprule
    Hyperparameter & Value \\
    \midrule
    Rank ($r$)           & 8    \\
    Alpha ($\alpha$)     & 16   \\
    Dropout              & 0.05 \\
    Bias                 & none \\
    Task type            & Causal LM \\
    \bottomrule
  \end{tabular}
  \caption{LoRA hyperparameters shared across all fine-tuning runs.}
  \label{tab:lora_hyperparams}
\end{table}

\section{Additional Results}
\label{sec:additional_results}
This appendix reports results on SmolLM3-3B \cite{bakouch2025} and Qwen2.5-1.5B \cite{qwen2025}, to assess whether the trends observed in the main experiments persist across architectures and parameter scales.
Overall, the same pattern holds: (i) domain fine-tuning is crucial (all fine-tuned variants strongly outperform the pretrained baselines), and (ii) the proposed Topological loss provides the most consistent improvements on recipe-specific factual and procedural metrics, particularly ingredient recall, quantity precision, and action/step edit distances.

\subsection{SmolLM3-3B}
\Cref{tab:results_smollm3} confirms the results when changing architecture (and pretraining data), showing that CE fine-tuning already yields significant gains over the pre-trained model across all metrics, confirming the necessity of fine-tuning to learn the recipe structure and cooking-action patterns.
Among the composite objectives, Topological achieves the best overall balance: it yields the highest ROUGE-1 and BERTScore, improves Action Precision, Quantity Precision, and Ingredient Recall over CE, and reduces both Action and Step edit distances (better procedural alignment). Dice remains competitive and provides the strongest Temperature Precision on this backbone, suggesting that set-level overlap objectives help reliably recover critical numeric tokens, while the topological term primarily strengthens ingredient-set semantic coherence.

\begin{table*}[htb]
\sisetup{table-format=3.2, round-precision=2, round-mode=places, detect-all}
\centering
\begin{tabular}{l |S S| S S S S S S S}
\toprule
Model & {R1 $\uparrow$} & {BS $\uparrow$} & {AP $\uparrow$} & {QP $\uparrow$} & {IR $\uparrow$} & {TeP $\uparrow$} & {TiP $\uparrow$} & {AD $\downarrow$} & {SD $\downarrow$} \\ \midrule
PreTrained & 16.0909 & 86.4398 & 31.0831 & 22.2941 & 22.0935 & 37.5040 & 37.3020 & 48.4039 & 56.0939 \\ \midrule
CE & 21.8935 & 88.7898 & 34.8390 & 47.7397 & 32.0394 & 56.6698 & 44.4984 & \bfseries 41.0928 & \bfseries 42.0928 \\
Focal & 20.3938 & 89.0018 & 34.4904 & 47.3939 & 31.0928 & 56.3903 & 44.3098 & 42.9830 & 43.8929 \\
Dice & \itshape 23.0849 & \itshape 89.5904 & \itshape 35.0967 & \itshape 49.5758 & \itshape 34.0976 & \bfseries 58.9575 & \bfseries 46.4484 & 43.5958 & 44.5958 \\
Topological & \bfseries 25.9079 & \bfseries 89.7789 & \bfseries 37.5597 & \bfseries 51.8961 & \bfseries 36.9899 & \itshape 57.9797 & \itshape 45.8889 & \itshape 41.5958 & \itshape 42.0973 \\ \bottomrule
\end{tabular}
\caption{Results for finetuned and pretrained SmolLM3-3B models using ROUGE-1 (R1), BERTScore (BS), Action Precision (AP), Quantity Precision (QP), Ingredient Recall (IR), Temperature Precision (TeP), Time Precision (TiP), Action Distance (AD), Step Distance (SD). In \textbf{bold} the top performance, and in \textit{italic} the second best.}
\label{tab:results_smollm3}
\end{table*}

\subsection{Qwen2.5-1.5B}
\Cref{tab:results_qwen25_15b} confirms the same overall behavior on a smaller 1.5B backbone: fine-tuning substantially improves not only surface-form metrics (ROUGE-1) but also executability-oriented measures (ingredient recall, quantity precision, and lower procedural edit distances).
The Topological objective again provides the greatest improvements on ingredient- and procedure-related metrics, achieving the best ROUGE-1, BERTScore, Ingredient Recall, and the lowest Action/Step edit distances.
Dice remains a strong alternative when prioritizing numerical faithfulness, attaining the best Temperature Precision and Time Precision in this setting, which supports the view that Dice and Topological objectives emphasize complementary aspects of recipe correctness.

\begin{table*}[htb]
\sisetup{table-format=3.2, round-precision=2, round-mode=places, detect-all}
\centering
\begin{tabular}{l |S S| S S S S S S S}
\toprule
Model & {R1 $\uparrow$} & {BS $\uparrow$} & {AP $\uparrow$} & {QP $\uparrow$} & {IR $\uparrow$} & {TeP $\uparrow$} & {TiP $\uparrow$} & {AD $\downarrow$} & {SD $\downarrow$} \\ \midrule
PreTrained & 9.49 & 85.42 & 27.40 & 18.89 & 14.88 & 31.10 & 33.09 & 57.99 & 63.45 \\ \midrule
CE & 14.10 & 85.10 & 27.87 & 41.76 & 26.48 & 47.79 & 39.90 & 47.97 & 55.56 \\
Focal & 13.99 & 85.25 & 26.79 & 41.49 & 25.32 & 47.00 & 39.33 & 50.86 & 58.97 \\
Dice & 14.38 & 85.50 & 28.61 & 42.68 & 27.49 & \bfseries 48.70 & \bfseries 42.91 & 46.49 & 52.67 \\
Topological & \bfseries 16.86 & \bfseries 87.79 & \bfseries 29.82 & \bfseries 43.50 & \bfseries 29.29 & 47.08 & 40.91 & \bfseries 45.39 & \bfseries 51.98 \\ \bottomrule
\end{tabular}
\caption{Results for finetuned and pretrained Qwen2-1.5B models using ROUGE-1 (R1), BERTScore (BS), Action Precision (AP), Quantity Precision (QP), Ingredient Recall (IR), Temperature Precision (TeP), Time Precision (TiP), Action Distance (AD), Step Distance (SD). In \textbf{bold} the top performance, and in \textit{italic} the second best.}
\label{tab:results_qwen25_15b}
\end{table*}

\section{Human Evaluation Details}
\label{sec:human_evaluation_details}
This section provides additional details on the human evaluation and its settings.

\subsection{Participants and recruitment.}
Three adult volunteers were recruited via convenience sampling from the authors' personal network: (i) one hobbyist home cook, (ii) one professional pastry chef, and (iii) one participant with professional experience as a baker and pizza maker with restaurant cooking experience. All participants were located in Italy and evaluated the outputs in English as presented.
Participation was unpaid and fully voluntary, and participants could stop at any time without providing a reason.

\subsection{Human Evaluation Instructions}
\label{sec:human_eval_instructions}

\paragraph{Materials.}
For each evaluated prompt, we generated two structured recipes (ingredients + steps) using (a) CE fine-tuning and (b) Topo+Dice fine-tuning, using identical decoding settings. To reduce bias, the two candidates were randomized in order and labeled as \textit{Recipe A/B}.

\paragraph{Purpose.}
You will compare two automatically generated structured recipes for the same cooking prompt. The goal is to help identify which recipe is more plausible and to spot concrete errors.

\paragraph{What you will see.}
For each prompt, you will see:
(1) the recipe name (e.g., ``Pasta Carbonara'');
(2) two candidate recipes labeled \textbf{Recipe A} and \textbf{Recipe B}, each with an ingredients list and step-by-step instructions.

\paragraph{Your tasks (for each prompt).}
\begin{enumerate}
    \item \textbf{Preference:} Choose which recipe you prefer (A or B), assuming a home cook will execute it based on:
     \begin{itemize}
         \item the ingredient completeness and plausibility
         \item the plausibility of numbers (times, temperatures, and quantities)
         \item the procedural plausibility and clarity
         \item the overall quality of the recipe
     \end{itemize}
    \item \textbf{Error spotting:} Mark any problems you notice, using any of these categories:
    \begin{itemize}
        \item Ingredient issues: missing essential ingredients, incompatible combinations, clearly wrong substitutions.
        \item Quantity issues: unrealistic amounts, inconsistent units, quantities that contradict steps.
        \item Time issues: unrealistic or unsafe values.
        \item Temperature issues: unrealistic or unsafe values.
        \item Procedural issues: wrong ordering, missing crucial steps, contradictions.
        \item Hallucination: absurd steps or ingredients.
        \item Safety Issues: unsafe procedure or ingredients that could cause harm if the recipe is cooked.
    \end{itemize}
    \item \textbf{Notes:} Eventual notes not included in the previous evaluation.
\end{enumerate}

\paragraph{Risks and notes.}
This is not a medical or professional consultation on food safety. Some generated recipes may contain mistakes (including potentially unsafe suggestions). Please do not cook from these recipes as they are.
Stop at any time if you feel uncomfortable.

\subsection{Consent Form}
\label{sec:consent}
Before starting, each participant received a short written information sheet and provided explicit consent to participate.
We recorded only participants' preference choices and free-text feedback on model outputs, along with coarse background descriptors relevant to the evaluation (i.e, home cook, pastry chef, baker-pizza maker; country: Italy).
In the ARR submission, participants are not identified by name.
With participants' explicit permission, we may acknowledge them by name in the camera-ready version; otherwise, acknowledgements will remain anonymous.
No institutional ethics review board approval was sought for this study because it consisted of an unpaid, minimal-risk, voluntary feedback activity with adults and did not involve the collection of sensitive personal data. Participants were informed of the study's purpose and their right to withdraw at any time.

\paragraph{Study title.}
Qualitative evaluation of automatically generated cooking recipes.

\paragraph{What you will do.}
You will review pairs of generated recipes and (i) choose which you prefer and (ii) mark errors you notice. The activity takes approximately 5 hours in total.

\paragraph{Voluntary participation.}
Your participation is voluntary. You may skip any question and may stop participating at any time without any consequences.

\paragraph{Compensation.}
There is no payment for participation.

\paragraph{Data collected and privacy.}
We will record your written feedback and your preference choices. We will also record a coarse description of your cooking background. We will not record your name or any identifying information in the paper. Results will be reported in aggregate or with anonymized quotes.

\paragraph{Risks.}
Minimal risk. Some recipes may include incorrect or unsafe cooking instructions; please do not execute the recipes as written.

\paragraph{Consent statement.}
By proceeding, you confirm that you are at least 18 years old, that you have read the information above, and that you consent to participate in this study.

\paragraph{Optional: being named in the paper.}
You may optionally allow us to include your name in the acknowledgements section of the final (camera-ready) paper.
This is optional and not required to participate.

\paragraph{Permission (choose one).}
\begin{itemize}
    \item I DO give permission to include my name in the acknowledgements.
    \item I DO NOT give permission to include my name in the acknowledgements.
\end{itemize}

\subsection{Human Evaluation Statistics}
This subsection reports detailed statistics from the human evaluation study, including inter-annotator agreement, preference distributions, and error analysis.

\subsubsection{Agreement and Preference Distribution}
\Cref{tab:human_preferences} shows preference statistics across all evaluation dimensions. Inter-annotator agreement was measured using Fleiss' $\kappa$, which quantifies the degree of agreement beyond chance among multiple raters evaluating categorical items. All dimensions achieved substantial agreement according to the interpretation guidelines of \citet{Landis1977}: ingredients ($\kappa = 0.69$), numbers ($\kappa = 0.69$), procedure ($\kappa = 0.67$), and overall quality ($\kappa = 0.75$), with a mean of $\kappa = 0.70$ . At the recipe level, 48 out of 64 recipes (75\%) achieved unanimous agreement among all three annotators on overall preference, while 16 recipes (25\%) had majority agreement (2 out of 3 annotators).

The raw preference percentages indicate that Topo+Dice was consistently preferred across all dimensions, with overall quality showing the strongest preference (62\% for Topo+Dice vs. 11\% for CE, with 27\% tied). When excluding tie cases (which is standard practice for computing win rates in pairwise comparisons), Topo+Dice achieved win rates ranging from 79.3\% (ingredients) to 88.0\% (procedure), with an overall win rate of 84.3\%. The high proportion of ties in certain dimensions (52\% for ingredients, 49\% for numbers) indicates that many generated recipes were of comparable quality on specific aspects, but when annotators could distinguish quality differences, they favored Topo+Dice. Statistical significance was confirmed using both $\chi^2$ tests and McNemar's test, with all comparisons yielding $p < 0.01$.

\begin{table}[htb]
    \centering
    \resizebox{\linewidth}{!}{
        \begin{tabular}{l|cc|ccc|c}
            \toprule
            \textbf{Dimension} & \textbf{$\kappa$} & \textbf{Agr.} & \textbf{TD} & \textbf{Tie} & \textbf{CE} & \textbf{Win} \\\midrule
            Ingredients & .69 & .72 & .38 & .52 & .10 & .79 \\
            Numbers & .69 & .70 & .42 & .49 & .9 & .81 \\
            Procedure & .67 & .69 & .46 & .48 & .6 & .88 \\
            Overall & .75 & .75 & .62 & .27 & .11 & .84 \\\bottomrule
        \end{tabular}
    }
    \caption{Human evaluation results across 192 judgments (64 recipes × 3 annotators). 
    $\kappa$: Fleiss' $\kappa$; Agr.: full agreement (3/3); TD/CE: Topo+Dice/CE preference; 
    Win\%: TD win rate excluding ties. All differences are significant at p<0.01 according to $\chi^2$ and McNemar tests.}
    \label{tab:human_preferences}
\end{table}

\subsubsection{Error Analysis}
\Cref{tab:error_analysis} details the errors identified by annotators across 192 evaluations. We find a dramatic reduction in critical structural errors: step mismatches decreased from 14 (CE) to 2 (Topo+Dice), missing ingredients from 14 to 3, incorrect quantities from 13 to 2, and hallucinations from 4 to 0. These reductions (ranging from 78.6\% to 100\%) demonstrate that the topological loss effectively addresses the coherence and factual correctness issues that affect standard cross-entropy training.
The "Other" category encompasses various issues, including time inconsistencies (7 cases for Topo+Dice, 3 for CE), temperature discrepancies (1 vs. 3), and one safety issue in CE. The relative increase in time-related errors for Topo+Dice (7 vs 3) suggests that while the model excels at structural and ingredient coherence, temporal specifications remain an area for future improvement. However, these are less severe than the structural errors eliminated by our approach. In total, annotators flagged errors in only 7.8\% of Topo+Dice evaluations compared to 24.0\% for CE, representing an overall error reduction of 67.5\%. This substantial improvement in generation quality aligns with the preference results and validates the effectiveness of incorporating geometric structure into the training objective.

\begin{table}[htb]
    \centering
    \begin{tabular}{l|cc|c}
        \toprule
        Error Type & TD & CE & Reduction \\\midrule
        Step Mismatch & 2 & 14 & 85.7\% \\
        Missing Ingredient & 3 & 14 & 78.6\% \\
        Bad Quantity & 2 & 13 & 84.6\% \\
        Hallucination & 0 & 4 & 100\% \\
        Other & 8 & 7 & -- \\\midrule
        \textbf{Total Errors} & \textbf{15} & \textbf{46} & \textbf{67.5\%} \\\bottomrule
    \end{tabular}
    \caption{Error analysis across 192 evaluations. Topo+Dice (TD) eliminates 
    critical structural errors (step mismatches, missing ingredients, hallucinations) 
    that affect the CE baseline.}
    \label{tab:error_analysis}
\end{table}

\section{NER Validation}
To assess the reliability of the Qwen3-4B-based NER pipeline, we conducted a post hoc validation study on a random subset of 60 generated recipes. We re-ran entity extraction with Gemini 2.0 as an independent, stronger reference system and further verified both extractors against human annotations on 20 of those samples. 

\Cref{tab:ner_f1} reports pairwise F1 scores measuring agreement between (i) Qwen and Gemini and (ii) Gemini and human annotators across all five entity types. Both comparisons provide high F1 values (>91\%), with Gemini–Human agreement consistently at or above 94\%, confirming that Qwen-extracted entities closely approximate human-level annotations, such as Gemini ones.

\begin{table}[htb]
    \centering
    \begin{tabular}{@{}l|cc@{}}
    \toprule
                 & Qwen-Gemini    & Gemini-Human    \\ \midrule
    Quantity     & 97.06             & 99.43              \\
    Temperature  & 91.67             & 99.67              \\
    Time         & 93.13             & 94.44              \\
    Ingredients  & 97.06             & 99.43              \\
    Instructions & 97.35             & 99.33              \\ \bottomrule
    \end{tabular}
    \caption{NER F1 scores (on 60 recipes) between the Qwen3-4B and Gemini 2.0 extractors (Qwen-Gemini) and between Gemini and human annotations on a 20-recipe sub-sample (Gemini-Human) across the five entity types.}
    \label{tab:ner_f1}
\end{table}

\Cref{tab:gemini_metrics} reports the task metrics recomputed on the 60-sample subset using Gemini-extracted entities, together with the absolute shift ($\Delta$) relative to the Qwen-extracted scores for the CE and Topo+Dice (TD) configurations.
The shifts are near-uniform ($\approx+1$ point for all metrics, except $\approx+2.5$ points for step distance) and nearly identical between CE and TD, confirming two key properties: (i) the extraction noise introduces only a small, consistent offset that does not distort any individual metric, and (ii) the offset is non-differential. Since it affects all configurations equally, the relative 
ranking across all loss objectives is fully preserved regardless of which extractor is used.

\begin{table}[htb]
    \centering
    \begin{tabular}{@{}l|cc@{}}
    \toprule
    Metric & CE ($\Delta$) & TD ($\Delta$) \\ \midrule
    QP     & 50.64 (+0.97) & 65.66 (+1.02) \\
    IR     & 37.20 (+1.00) & 47.75 (+1.04) \\
    AP     & 45.21 (+1.03) & 56.39 (+0.96) \\
    TeP    & 62.68 (+0.97) & 67.64 (+0.97) \\
    TiP    & 53.13 (+0.98) & 63.19 (+0.99) \\
    AD     & 30.68 (+0.83) & 30.69 (+0.88) \\
    SD     & 32.48 (+2.53) & 32.50 (+2.51) \\\bottomrule
    \end{tabular}
    \caption{Task metrics recomputed on the 60-recipe validation subset using Gemini 2.0 as the NER extractor for CE and TD. $\Delta$ denotes the absolute shift from the corresponding Qwen-extracted scores.}
    \label{tab:gemini_metrics}
\end{table}

\section{Computational Overhead}
Inference-time complexity is identical across all objectives, since the model architecture and decoding procedure are unchanged. The proposed losses affect training only, with a modest overhead as shown in \Cref{tab:computation_cost}. Compared to the CE baseline, the additional wall-clock training time is at most 4 minutes (+6.7\%), while the topological loss gets the largest GPU utilization increase of +9.61 percentage points as expected. Given these marginal costs, all custom objectives remain practical for standard single-GPU fine-tuning runs.

\begin{table}[htb]
    \centering
    \begin{tabular}{@{}l|cc@{}}
    \toprule
    Loss  & Training time & Avg GPU util. \\ \midrule
    CE    & 59 min        & 59.74\%       \\
    Focal & 60 min        & 61.78\%       \\
    Dice  & 62 min        & 63.12\%       \\
    Topo  & 63 min        & 69.35\%       \\ \bottomrule
    \end{tabular}
    \caption{Training time (wall-clock, in minutes) and average GPU utilization of each loss objective during fine-tuning, measured on a single GPU using}
    \label{tab:computation_cost}
\end{table}

\end{document}